\documentclass[letterpaper, 10 pt, conference]{ieeeconf}  

\IEEEoverridecommandlockouts                              

\usepackage{bm}
\usepackage{url}
\usepackage{graphicx}
\usepackage{lipsum}
\usepackage{pifont}
\usepackage{amsmath}
\usepackage{amssymb}
\usepackage{nccmath}
\usepackage{comment}
\usepackage{balance}
\usepackage{multirow}
\usepackage{textcomp}
\usepackage{multirow}
\usepackage{subcaption}
\usepackage{booktabs}
\usepackage{algorithm}
\usepackage{dblfloatfix}
\usepackage[table]{xcolor}
\usepackage[noend]{algpseudocode}

\title{\LARGE \bf
3D-BBS: Global Localization for 3D Point Cloud  \\ Scan Matching Using Branch-and-Bound Algorithm
}

\author{Koki Aoki$^{1}$, Kenji Koide$^{2}$, Shuji Oishi$^{2}$, Masashi Yokozuka$^{2}$, Atsuhiko Banno$^{2}$, and Junichi Meguro$^{1}$
\thanks{*This work was supported in part by a project commissioned by the New Energy and Industrial Technology Development Organization (NEDO).}%
\thanks{$^{1}$Koki Aoki and Junichi Meguro are with the Department of Mechatronics Engineering, Meijo University, 1-501 Shiogamaguchi, Tempaku-ku, Nagoya 4688502, Japan, {\tt\small  223432001@ccmailg.meijo-u.ac.jp}}
\thanks{$^{2}$Kenji Koide, Shuji Oishi, Masashi Yokozuka, and Atsuhiko Banno are with the Department of Information Technology and Human Factors, the National Institute of Advanced Industrial Science and Technology, Umezono 1-1-1, Tsukuba, 3050061, Ibaraki, Japan.}
}
\begin{document}

\definecolor{verylightgray}{gray}{0.95}
\newcolumntype{g}{>{\columncolor{verylightgray}}c}

\maketitle
\thispagestyle{empty}
\pagestyle{empty}

\setlength\floatsep{8pt}
\setlength\textfloatsep{10pt}

\newcommand{\argmax}{\mathop{\rm argmax}\limits}
\newcommand{\argmin}{\mathop{\rm argmin}\limits}
\begin{abstract}
This paper presents an accurate and fast 3D global localization method, 3D-BBS, that extends the existing branch-and-bound (BnB)-based 2D scan matching (BBS) algorithm. To reduce memory consumption, we utilize a sparse hash table for storing hierarchical 3D voxel maps. To improve the processing cost of BBS in 3D space, we propose an efficient roto-translational space branching. Furthermore, we devise a batched BnB algorithm to fully leverage GPU parallel processing. Through experiments in simulated and real environments, we demonstrated that the 3D-BBS enabled accurate global localization with only a 3D LiDAR scan roughly aligned in the gravity direction and a 3D pre-built map. This method required only 878 msec on average to perform global localization and outperformed state-of-the-art global registration methods in terms of accuracy and processing speed.
\end{abstract}
\section{Introduction}
3D light detection and ranging sensor (LiDAR)-based localization has played an important role in the mobile robot. LiDAR-based global localization, which is the task of finding a robot's global pose in a pre-built map without any initial estimates, is effective in dense building environments where the global navigation satellite systems (GNSS) accuracy decreases due to the multipath and signal blockage.
This technique is a key technology in robotics applications such as initial pose estimation. However, LiDAR-based global localization with a large map in real time is challenging because of the large search space.

There are several approaches to solving the global localization problem. Global registration methods \cite{teaser}, \cite{quatro}, \cite{fgr} extract point features to describe local geometrical shapes and relationships. A transformation is then estimated by establishing point-to-point correspondences. Although these methods can robustly match a pair of 3D point clouds with outliers \cite{teaser}, they require considerable processing time for point correspondence estimation with a large map point cloud and often suffer from repeated environmental structures.

For scan-to-map matching, Hess et al. \cite{hess} proposed a 2D global localization method based on the branch-and-bound (BnB) algorithm. Their method used hierarchical occupancy grid maps to compute the upper bound of scan matching scores and efficiently prune unpromising candidate poses through hierarchical search space tree branching. This approach enables precise and fast global localization in 2D maps. However, these performances are difficult to achieve in 3D maps because memory consumption and processing costs drastically increase.

In this work, we propose a 3D global localization method using BnB-based scan matching (3D-BBS). 
Our method estimates the sensor pose on a 3D pre-built map with a single 3D LiDAR scan roughly aligned in the gravity direction. 3D-BBS reduces memory consumption and processing time using sparse map structures and GPU-accelerated batched BnB algorithm. In our experiments, 3D-BBS exhibits accurate and fast global localization in a second on average, and significantly outperforms the state-of-the-art global registration methods~\cite{teaser}, \cite{quatro}, \cite{fgr}.

\begin{figure}[t]
  \centering
  \includegraphics[width=1.0\linewidth]{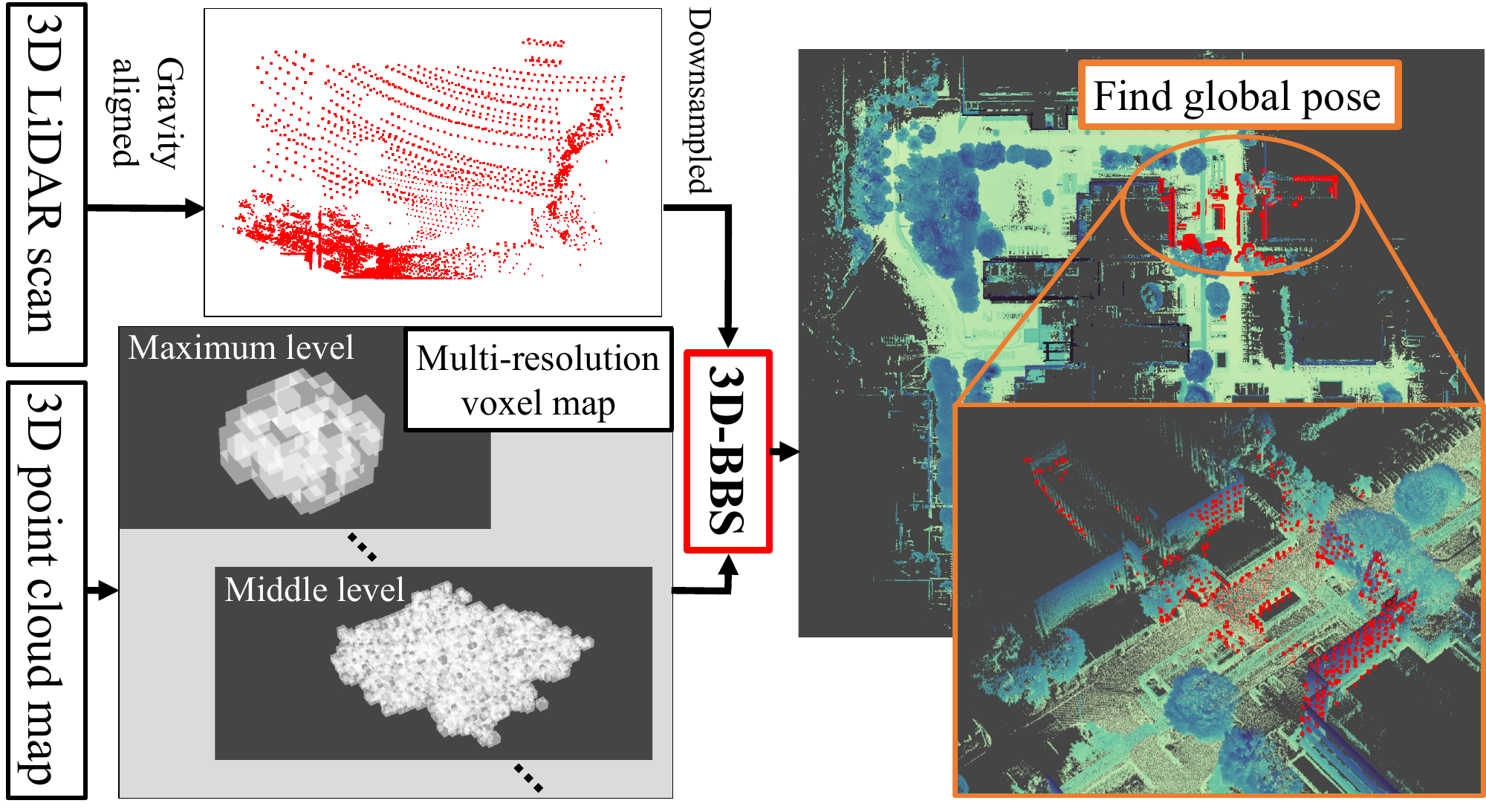}
  \caption{Overview of the proposed global localization.}
  \label{fig:overview}
\end{figure}

The main contributions of this work are as follows:
\begin{itemize}
  \item We propose a batched BnB algorithm that quickly computes a large number of score calculations at once on the GPU.
  \item For a faster BnB algorithm, we employ a combination of the best-first search and roto-translational branching. We empirically confirmed that it can drastically reduce the processing time with a slight approximation.
  \item We utilize a 3D voxel map created with a sparse hash table that enables saving the memory required for a 3D multi-resolution voxel map. The code of the proposed algorithm is available on a GitHub repository\footnote{\url{https://github.com/KOKIAOKI/3d_bbs}}.
\end{itemize}
\section{Related Work}
\subsection{Global Localization with Point Cloud Matching}
For global pose estimation, many studies have utilized feature extraction of point clouds \cite{gl_survey}. In this approach, descriptors that represent local geometrical shapes, such as FPFH \cite{fpfh}, and SHOT \cite{shot} are extracted to determine the point correspondences between point clouds. Subsequently, the relative pose between the point clouds is estimated using a robust pose estimator such as RANSAC \cite{ransac}. TEASER~\cite{teaser} utilizes a graph-theoretic consensus finding algorithm that enables robust pose estimation for a large number of outliers. Although these methods achieve fast alignment in scan-to-scan global registration, the registration of scan to large map would require large processing time to find the correspondence.

Chen et al. \cite{range-image} use the range images generated from LiDAR scans and a meshed map. Their method achieved global localization for various datasets using an observation model for a Monte Carlo localization (MCL) \cite{mcl}. However, a large number of particles are required to initialize accurately, and convergence takes a long time.

Frame-based methods extract geometrical features to represent a whole point cloud frame and identify revisited places by finding frames with similar descriptors. 
Scan context~\cite{scancontext} creates a database of descriptors based on the height of 3D LiDAR points and effectively detects the loops in a trajectory. 
OverlapNet \cite{overlapnet} utilizes a deep neural network exploiting different cues generated from LiDAR data, and estimates the overlap rates and yaw angle between two point clouds. 
OverlapTransformer \cite{overlaptransformer} uses only LiDAR data, and achieves loop closure candidates detection faster than OverlapNet.
BoW3D \cite{bow3d} built a dictionary of bag-of-wards based on Link3D \cite{link3d} features representing LiDAR scan keypoints and achieved real-time performance in 6DoF loop closure. 
While frame-based methods require keyframes of LiDAR scans that cover the entire map, 3D-BBS directly uses 3D maps that have already been constructed.

\subsection{Point Cloud Matching Using Branch-and-Bound Method}
Several studies have used the BnB algorithm \cite{first-bnb}, \cite{bnb-survey} to quickly estimate the accurate global pose. Because the BnB algorithm is a full search algorithm, a pose with the global best matching cost can be estimated more reliably than with methods using Monte Carlo algorithms. Olsson et al. \cite{Olsson} propose a BnB-based registration algorithm that finds the global optima of the non-convex registration problems. Go-ICP \cite{goicp} combines the iterative closest point (ICP) \cite{icp} with the BnB algorithm to find the globally optimal solution. However, these methods require an initial pose estimate. Hess et al. \cite{hess} develop BnB-based 2D scan matching by using hierarchical occupancy grid map sizes. Although this method enables accurate and fast global localization on 2D maps, the processing time increases drastically for 3D maps.
\section{Methodology}
\subsection{Problem Setting}
Inspired by \cite{hess}, we propose a global localization framework, 3D-BBS, as shown in Fig. \ref{fig:overview}. The 3D global localization problem is defined as the task of finding the 6DoF sensor pose $\boldsymbol{x}=[x, y, z,\alpha, \beta, \gamma]^\top$, where $(x, y, z)$ represents the translation, and $(\alpha, \beta, \gamma)$ denotes Euler angles corresponding to roll, pitch, and yaw, respectively. Let $\bm{T}_{\bm{x}}$ be the transformation matrix of $\bm{x}$. Given a 3D point cloud map $\mathcal{M}=\{\bm{m}_n \in\mathbb{R}^3 | _{n=1,\ldots, N}\}$ and 3D LiDAR scan points $\mathcal{S}=\{\bm{s}_k\in\mathbb{R}^3 |_{k=1, \dots, K}\}$, 3D-BBS estimates the sensor pose~$\boldsymbol{x}^{*}$ on the map.
We assume that the gravity direction of $\mathcal{S}$ is roughly estimated, for example, by using an inertial measurement unit (IMU), and perform scan matching in full 4DoF (XYZ translation and yaw angle) with a small roll and pitch search range.

We review the outline of the 2D BBS technique in Sec.~\ref{sec:2dbbs}, and then extend it to 3D in Sec. ~\ref{sec:voxelmap} to \ref{sec:batch}.
\subsection{2D Branch-and-Bound Scan Matching}
\label{sec:2dbbs}
Hess et al. \cite{hess} proposed branch-and-bound scan matching~(BBS) in the context of global localization for 2D SLAM loop detection. Let $M:\mathbb{R}^2 \to \mathbb{Z}$ be an occupancy grid map from a point on the grid into the occupancy value of the nearest grid with cell size $r$. Given 2D LiDAR scan points $\mathcal{S}=\{\bm{s}_k\in\mathbb{R}^2 |_{k=1, \dots, K}\}$, the BBS estimates the global solution of the 3DoF sensor pose $\bm{x}^{*}$ as follows:
\begin{align}
\label{2dbbs}
\bm{x}^{*} = \argmax_{\substack{
x \in\{ ri_x | i_x \in\mathbb{Z}, 
W_{\text{min}}^x\le ri_x \le W_{\text{max}}^x \} \\ 
y \in\{ ri_y | i_y \in\mathbb{Z}, 
W_{\text{min}}^y\le ri_y \le W_{\text{max}}^y \} \\ 
\theta \in\{ \delta(r)i_{\theta} | i_{\theta} \in\mathbb{Z}, 
W_{\text{min}}^{\theta} \le \delta(r)i_{\theta} \le W_{\text{max}}^{\theta} \} } } 
\quad \sum_{k=1}^K M(\bm{T}_{\bm{x}} \bm{s}_k),
\end{align}
where $K$ is the number of the input LiDAR scan, $\bm{T}_{\bm{x}}$ is the 3 DoF transformation matrix of pose hypothesis $\bm{x} = [x,y,\theta]^{\top}$, and
$(W_{\text{min}}^x,W_{\text{max}}^x)$, $(W_{\text{min}}^y,W_{\text{max}}^y)$,
$(W_{\text{min}}^{\theta},W_{\text{max}}^{\theta})$ are translational and rotational search ranges, respectively.

{\bf Branch-and-bound algorithm:} The BnB algorithm, which is a key component of BBS, uses a hierarchical tree search to quickly find a solution that is equivalent to the result of the full search. The tree structure and leaf nodes represent the search space and feasible solutions to the problem, respectively. Nodes in an upper level represent a set of child nodes (i.e., branching). At each non-leaf node, an upper bound estimate of the scores of its children is calculated. If the calculated upper bound is lower than the provisional best score, we can prune that node with children nodes without approximation (i.e., pruning). If a high provisional best score is found earlier, the BnB algorithm can prune many nodes, resulting in a large reduction of the computation cost. Therefore, the order of the search is important to maximize the performance of the BnB algorithm (i.e., search strategy). 

{\bf Branch-and-bound scan matching \cite{hess}:} To solve the problem of 2D global localization using the BnB algorithm, the pose space within the search range is represented as a tree structure. Let $l \in\mathbb{Z}, 0 \le l \le l_{\text{max}}$ be the hierarchy level in the tree, where $l_{\text{max}}$ is the maximum level. The tree is branched from large to small step sizes. The step size at each level is determined based on the cell size $r$ of the occupancy grid map. 1) The   step size is $r_l=2^{l}r$. 2)~The rotational step size $\delta(r)$ is common for each level and selected such that a point in the maximum scan range $d_{\text{max}}$ does not move more than $r$ as follows:
\begin{align}
\label{delta}
  \delta(r) =\text{arccos}(1-\frac{r^2}{2d_{\text{max}}^2}).
\end{align}

Each node of the tree is represented by $c = (c_x, c_y, c_{\theta}, c_l)\in\mathbb{Z}^4$, where $c_x$, $c_y$, and $c_{\theta}$ are the discrete parameters representing the sensor pose $\bm{x}_c$, and $c_{l}$ is the level of the node $c$. 

The tree is branched from the maximum level $l_{\text{max}}$. An initial set $\mathcal{C}_{l_{\text{max}}}$ is created as the direct product set of the translational and rotational initial sets:
\begin{align}
\mathcal{C}_{l_{\text{max}}} &={\mathcal{C}}_{l_{\text{max}}}^x \times {\mathcal{C}}_{l_{\text{max}}}^y\times \mathcal{C}_{l_{\text{max}}}^{\theta} \times \{ l_{\text{max}} \},
\end{align} 
where the initial sets of each component are as follows: 
\begin{align}
w_{\text{min}}^x = \left\lfloor\frac{W_{\text{min}}^x}{2^{l_{\text{max}}}r}\right\rfloor, 
\quad w_{\text{max}}^x =  \left\lceil\frac{W_{\text{max}}^x}{2^{l_{\text{max}}}r}\right\rceil, \notag \\
w_{\text{min}}^y = \left\lfloor\frac{W_{\text{min}}^y}{2^{l_{\text{max}}}r}\right\rfloor, 
\quad w_{\text{max}}^y =  \left\lceil\frac{W_{\text{max}}^y}{2^{l_{\text{max}}}r}\right\rceil, \\
w_{\text{min}}^{\theta} = \left\lfloor\frac{W_{\text{min}}^{\theta}}{\delta(r)}\right\rfloor, 
\quad w_{\text{max}}^{\theta} =  \left\lceil\frac{W_{\text{max}}^{\theta}}{\delta(r)}\right\rceil, \notag \\
\mathcal{C}_{l_{\text{max}}}^x = \{ i_x \in \mathbb{Z} | w_{\text{min}}^x \le i_x \le w_{\text{max}}^x\}, \notag \\
\mathcal{C}_{l_{\text{max}}}^y= \{ i_y \in \mathbb{Z} | w_{\text{min}}^y \le i_y \le w_{\text{max}}^y\}, \\
\mathcal{C}_{l_{\text{max}}}^{\theta} = \{ i_{\theta} \in \mathbb{Z} | w_{\text{min}}^{\theta} \le i_{\theta} \le w_{\text{max}}^{\theta}\}.  \notag
\end{align} 

A non-leaf node $c$ branches into four children nodes as follows:
\begin{align}
\mathcal{C}_{c_l} = \{(2c_x +j_x, 2c_y + j_y, c_{\theta}, c_l -1) | (j_x,j_y)\in\{0,1\} \}.
\end{align}
At each branched node, the scan matching score is calculated as follows:
\begin{align}
\label{eq:2dbbs}
score(c) = \sum_{k=1}^K M(\bm{T}_{\bm{x}_c} \bm{s}_k), \\
\bm{x}_c = [r_{c_{l}} c_x,r_{c_{l}} c_y, \delta(r)c_{\theta}]^\top.
\end{align}
To calculate the upper bound without effort to find the maximum score of the children, $M_{\text{precomp}}^{l}$ with a cell size $2^lr$ is created at each level in advance. The upper bound can be efficiently computed as follows:
\begin{align}
\overline{score}(c) &= \sum_{k=1}^K M_{\text{precomp}}^{c_l}(\bm{T}_{\bm{x}_c} \bm{s}_k), \\
M_{\text{precomp}}^{l} (x,y) &= \max_{\substack{x' \in [x, x+r(2^{l}-1)] \\ y' \in [y, y+r(2^{l}-1)]}} M(x',y').
\end{align}

Algorithm \ref{algo:bbs} shows the outline of the 2D BBS algorithm. All nodes in $\mathcal{C}_{l_{\text{max}}}$ are initially saved to queue $C$ in the descending order of $score (c)$.
To quickly select a provisional best score, the conventional method uses a depth-first search (DFS), which first selects nodes at a lower level. Moreover, $best\_score$ is initialized by the score threshold $score\_threshold$ to ignore nodes that have poor matches.

Extending this algorithm to 3D introduces two challenges. First, multi-resolution 3D voxel maps consume a significant amount of memory. Second, the number of candidate pose nodes grows exponentially as the dimension increases. To overcome these problems, we propose the 3D-BBS algorithm with sparse voxel maps, rotational branching, and batch processing.
\begin{algorithm}[tb]
\caption{BBS \cite{hess}}
\label{algo:bbs}
\footnotesize
\begin{algorithmic}[1]
\State $best\_score$ $\gets$ $score\_threshold$
\While{$C$ is not empty}
  \State Pop $c$ from the queue $C$
  \If{$\overline{score}(c) > best\_score$}
    \If{$c$ is a leaf node}
      \State $match$ $\gets$ $c$
      \State $best\_score$ $\gets$ $\overline{score}(c)$
    \Else
      \State Branch: Split $c$ into nodes $\mathcal{C}_{c_l} $
      \State Compute and memorize a score for each element in $\mathcal{C}_{c_l} $
      \State Push $\mathcal{C}_{c_l} $ onto the queue $C$, sorted by depth level, 
      \Statex \hspace{4.2em} maximum score last.
    \EndIf
  \EndIf
\EndWhile
\State \textbf{return} $best\_score$ and $match$ when set.
\end{algorithmic}
\end{algorithm}
\subsection{Multi-resolution Voxel Map}
\label{sec:voxelmap}
To reduce the memory consumption for a 3D multi-resolution precomputed map, we introduce spatial hashing-based sparse voxel maps.
The voxel resolution at each level is expressed as $r_l = 2^{l}r$, where $r$ is the minimum voxel size. 
We create an array of hash buckets per level, and store 3D coordinate occupied voxels $\bm{v}=[v_x,v_y,v_z]^\top\in\mathbb{Z}^3, v_x =\lfloor x/r_l \rfloor,  v_y =\lfloor y/r_l \rfloor,  v_z =\lfloor z/r_l \rfloor$ to the hash bucket.
The function $H^l: \mathbb{R}^3 \to \{ 0,1\}$ is defined to search for the bucket storing $\bm{v} $ and return its binary occupancy.
To search a hash bucket, we utilize a spatial hashing function $hash(\bm{v})$ that computes the index of the corresponding hash bucket as follows:
\begin{align}
\label{eq:hash}
hash(\bm{v}) = f(\boldsymbol{v})  \; \text{mod} \; T_l,
\end{align}
where the function $f(\bm{v})$ computes a hash value from a tuple of integers \cite{hash}, and $T_l$ is the size of hash buckets at $l$. In hash buckets construction, we seek $T_l$ until the hash collision rate becomes small (e.g. 0.1\%).
To avoid hash collisions, we implement the open addressing method; if a bucket is already occupied by another voxel, we seek an empty bucket while incrementing the bucket index. 

For each resolution, the 3D voxel coordinates of all map points in $\mathcal{M}$ are calculated and stored in hash buckets.
For upper bound calculation, the voxels $\mathcal{V}_{\text{add}}^{l,\bm{v}}=\{[v_x-j_x,v_y-j_y,v_z-j_z]^\top | (j_x,j_y,j_z)\in\{{0, 1}\}\}$ adjacent to all occupied voxels are stored in hash buckets so that the upper bound estimate of a parent node exceeds the scores of its children.

Let $c = (c_x, c_y,c_z, c_{\alpha},c_{\beta},c_{\gamma}, c_l)\in\mathbb{Z}^7$ be a node with discrete parameters representing the 6DoF sensor pose $\bm{x}_c$. 
We compute the matching score of node $c$ that simultaneously maintains the upper bound of scores in its children by using the precomputed map as follows:
\begin{align}
\label{eq:3Dupperbound}
score(c) = \overline{score}(c) &=\sum_{k=1}^K H^{c_{l}} \left(  \bm{T}_{\bm{x}_{c}}\bm{s}_{k} \right).
\end{align}
\subsection{Branching Algorithm}
\begin{figure}[tb]
  \centering
  \includegraphics[width=1.0\linewidth]{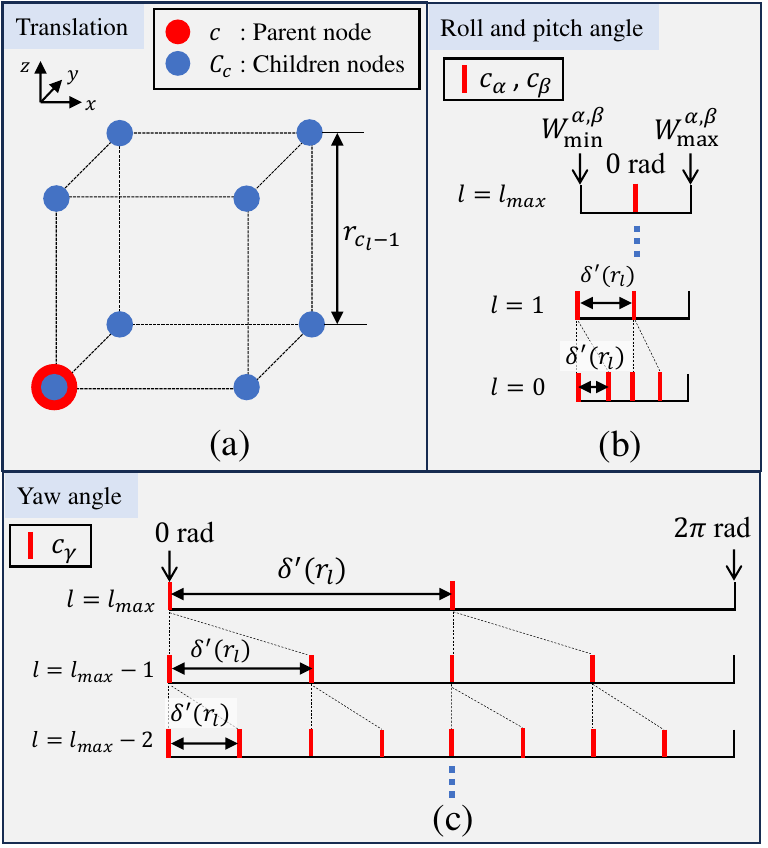}
  \caption{Overview of the branching operation in the proposed method. (a) Translational branching. (b) Roll and pitch angle branching. (c) Yaw angle branching.}
  \label{fig:branching-image}
\end{figure}
{\bf Branching rule:}
In contrast to [8], which only performs the branching of translational components, we perform both translational and rotational branching. The rotational branching makes the rotational step size sparse and drastically reduces the number of initial nodes. Fig. \ref{fig:branching-image} illustrates the proposed branching method. We have a translational search range equal to the bounding box of $\mathcal{M}$. In translational branching, a parent node branches into eight children nodes with step size $r_{{c_l}-1}$ (Fig.~\ref{fig:branching-image} (a)). The roll and pitch components are searched within a small range $(W_{\text{min}}^{\alpha,\beta}, W_{\text{max}}^{\alpha,\beta})$ (Fig.~\ref{fig:branching-image} (b)). The yaw angle search range is $(W_{\text{min}}^{\gamma}, W_{\text{max}}^{\gamma}) =(0,2\pi)$ (Fig.~\ref{fig:branching-image} (c)). In rotational branching, we choose a step size for each level such that it becomes smaller than $\delta(r)$ and divide the search range into segments of the same size as follows:
\begin{align}
 \delta'(r_l) = \frac{W_{\text{max}} - W_{\text{min}}}{\left\lceil\frac{W_{\text{max}} - W_{\text{min}}}{\delta(r_l)}\right\rceil}.
\end{align}
 The initial set $\mathcal{C}_{l_{\text{max}}}$ contains combinations of all poses within the search range at $l_{\text{max}}$. $\mathcal{C}_{l_{\text{max}}}$ is initially saved to queue $C$ in descending order of $score (c)$. Subsequently, a non-leaf node $c$ popped from $C$ is branched as follows:
\begin{align}
&\mathcal{C}_{c_l}  = \{ ( 2c_x+j_x, 2c_y+j_y, 2c_z+j_z, a_{\alpha}c_{\alpha}+j_{\alpha}, a_{\beta}c_{\beta}+j_{\beta}, \notag \\
 &a_{\gamma}c_{\gamma}+j_{\gamma}, c_{l}-1 ) | (j_x, j_y, j_z) \in \{0,1\}, j_{\alpha} \in \{ 0,...,a_{\alpha}-1 \}, \notag \\
 &j_{\beta} \in \{ 0,...,a_{\beta}-1 \}, j_{\gamma} \in \{ 0,...,a_{\gamma}-1 \} \} ,
\end{align}
where $a_{\alpha},a_{\beta}$ and $a_{\gamma}$ are the number of divisions for each rotational component of a node:
\begin{align}
  a=&\begin{cases}
    \left\lceil \frac{\delta'(r_{c_{l}})}{\delta'(r_{c_{l}-1})} \right\rceil & (\delta'(r_{c_{l}-1}) < (W_{\text{max}} - W_{\text{min}})) \\
   \: 1 & (\text{else}).
  \end{cases}
\end{align}
Note that by introducing angular branching, the upper bound estimate in Eq. \ref{eq:3Dupperbound} becomes not strictly exact. However, as shown in the experiments, we empirically confirmed that it underestimates the upper bound in only 0.001 \% of all cases.

{\bf Search strategy:}
Although we first remove nodes with estimated upper bounds less than $score\_threshold$ from the initial set, a large number of nodes remain in the queue. This causes a delay in finding the best solution for DFS and increases the processing time. Therefore, we employ BFS and sort $C$ in descending order of score. Because BFS selects a node with the best upper bound score, it tends to find a solution quicker than DFS when the global best score is close to the maximum score (i.e., when the LiDAR scan has a large overlap with the map). However,  in the middle levels, BFS requires branching a large number of high-score nodes because most of the input points overlap with inflated voxels.
To speed up the branching process, we implement GPU-parallelized batch processing for the score calculation.
\subsection{GPU-accelerated Batched Branch-and-Bound}
\label{sec:batch}
\begin{algorithm}[tb]
\caption{Batched 3D-BBS}
\label{algo:3d-bbs}
\footnotesize
\begin{algorithmic}[1]
\State $best\_score \gets score\_threshold$
\While{$C$ is not empty}
  \State Pop $c$ from the queue $C$
  \If{$\overline{score}(c) < best\_score$}
	\State \textbf{continue}
  \EndIf
  \If{$c$ is a leaf node}
    \State $match \gets c$
    \State $best\_score \gets \overline{score}(c)$
  \Else
      \State Branch: Split $c$ into nodes $\mathcal{C}_{c_l} $
      \State Add $\mathcal{C}_{c_l}$ to $\mathcal{C}_{\text{CPU}}$
  \EndIf
  \If {$|\mathcal{C}_{\text{CPU}}| > b$}
  \State $\mathcal{C}_{\text{GPU}} \gets \mathcal{C}_{\text{CPU}}$
  \State Compute and memorize a score for each element in $\mathcal{C}_{\text{GPU}}$
  \State $\mathcal{C}_{\text{CPU}} \gets \mathcal{C}_{\text{GPU}}$
  \State Push $\mathcal{C}_{\text{CPU}}$ onto the queue $C$, sorted by score
  \State Clear the all elements of $\mathcal{C}_{\text{CPU}}$.
  \EndIf
\EndWhile
\State \textbf{return} $best\_score$ and $match$ when set.
\end{algorithmic}
\end{algorithm}
We process the score calculations for multiple nodes in parallel on a GPU. To reduce the overhead of CPU-GPU data copy and synchronization, we implement batch processing. The procedure of our batched 3D-BBS is shown in Algorithm~\ref{algo:3d-bbs}. The branched nodes are stored in the node set $\mathcal{C}_\text{CPU}$ on the CPU memory. When the size of $\mathcal{C}_\text{CPU}$ exceeds the batch size $b$ (e.g., 10,000 nodes), $\mathcal{C}_\text{CPU}$ is transferred to the node set $\mathcal{C}_\text{GPU}$ on the GPU memory as a single memory block, and the scores of all nodes are calculated on the GPU. Subsequently, all the computed scores are sent back to $\mathcal{C}_\text{CPU}$. Note that voxel maps $H$ are copied to the GPU memory in advance to avoid run-time overheads.
\section{Experiments}
We evaluated the localization accuracy and processing time of the proposed method in simulated and real environments. To demonstrate the effectiveness of our algorithm, we used all combinations of processing types (CPU single-thread, CPU multi-thread (4 threads), and GPU), branching components (translation-only, translation + rotation), and search strategies (DFS and BFS) that resulted in nine configurations, as shown in Table \ref{tab:algorithms}. The combination of CPU single-thread, translation-only branching, and DFS (configuration (a)) corresponds to a naive 3D extension of \cite{hess}, and configuration (i) is our proposal.  In the real environment, we compared the proposed configuration with the state-of-the-art global registration methods TEASER++~\cite{teaser}, Quatro~\cite{quatro}, and FGR~\cite{fgr}.

To evaluate the estimation result, we used two criteria: 1) the translation error was smaller than 2.0 m, and 2) the rotation error was smaller than 0.05 rad. If a result satisfies both the translation and rotation criteria, we consider it sufficiently accurate to be used as an initial estimate for local fine registration \cite{gicp} to obtain a fine estimation result.

We used the same parameters for 3D-BBS through all the experiments. Considering the sizes of the experimental maps, we set $r=1.0$ m and $l_{\text{max}}=6$. Note that $r$ and $l_{\text{max}}$ should be changed according to the required accuracy and size of the map, respectively. To reduce the computational cost of the score calculation, all LiDAR scans were downsampled to approximately 1,000 points using a voxel grid filter. In the GPU implementation, batch size $b$ was empirically set to 10,000. The score threshold was set to 95 \rm{\%} of the total number of input points. The roll and pitch angle search range was respectively set to $(W_{\text{min}}^{\alpha,\beta}, W_{\text{max}}^{\alpha,\beta}) =(-0.02,0.02)$ rad to compensate for the gravity direction estimation errors.

We implemented the proposed algorithm in C++ and CUDA. All processes were performed with Intel Core i7-10700K 3.8Ghz 32GB and NVIDIA GeForce RTX2060.
\subsection{Simulated Environment}

\begin{table}[tb]
\centering
\caption{BnB algorithm configurations}
\label{tab:algorithms}
    \begin{tabular}{g|ggg}
        \toprule
        \rowcolor{white} & Processing & Branching& Search \\
	   \rowcolor{white} \multirow{-2}{*}{Configuration} & type &  components&srategy \\
        \midrule
          \rowcolor{white} (a) (ext. \cite{hess}) & CPU single-thread& Trans. & DFS \\
         (b) & CPU multi-thread & Trans. & DFS\\
          \rowcolor{white} (c) & GPU & Trans. & DFS \\
         (d) & CPU single-thread& Trans. & BFS \\
          \rowcolor{white} (e) & CPU multi-thread & Trans. & BFS  \\
         (f) & GPU & Trans. & BFS  \\
          \rowcolor{white} (g) & CPU single-thread& Trans. and rot. & BFS  \\
         (h) & CPU multi-thread & Trans. and rot. & BFS \\
         \rowcolor{white} (i) (ours) & GPU & Trans. and rot. & BFS  \\
        \bottomrule
    \end{tabular}
\end{table}

\begin{figure}[tb]
  \centering
  \includegraphics[width=0.6\linewidth]{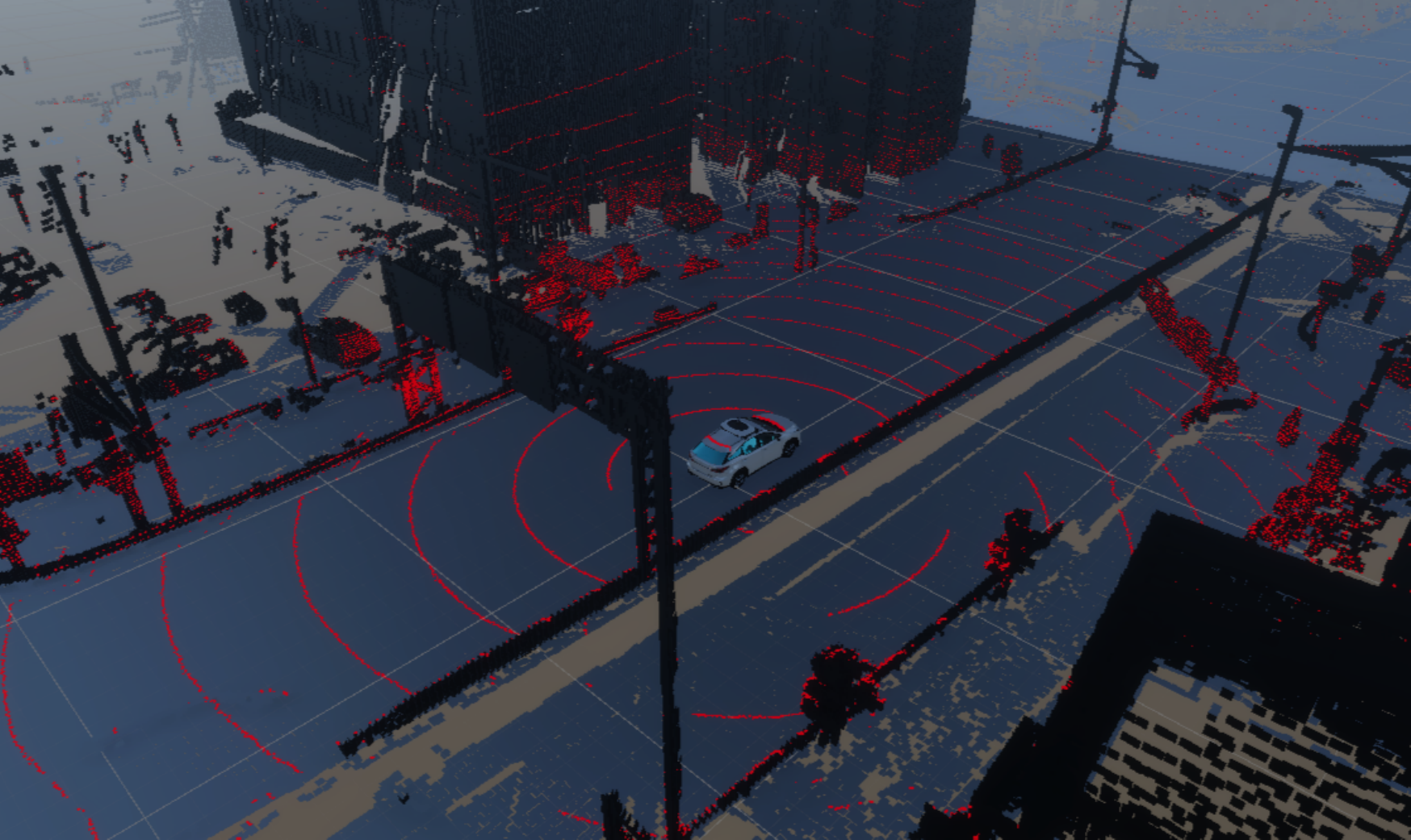}
  \caption{Simulated environment.}
  \label{fig:awsim}
\end{figure}

\begin{figure}[t]
  \centering
  \includegraphics[width=1.0\linewidth]{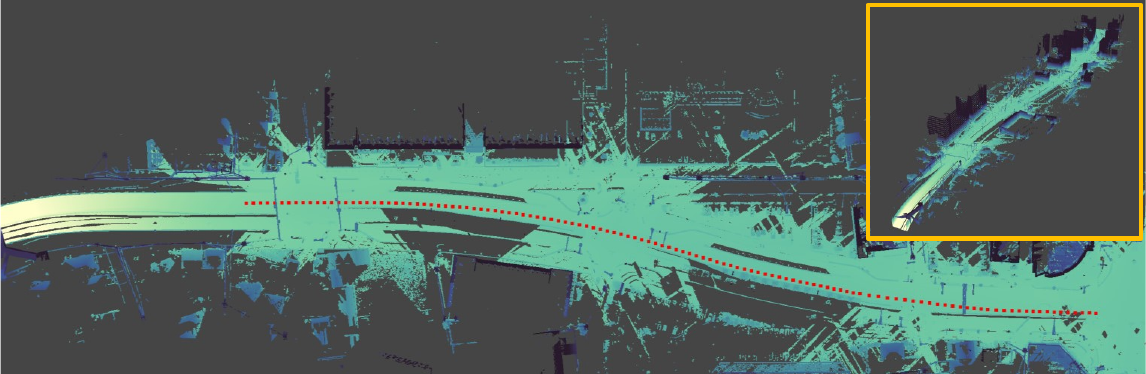}
  \caption{3D point cloud map in the experiment with the simulated environment. The map size is 211.0 $\times$ 729.7 $\times$ 68.5 $\rm{m^{3}}$. The total number of evaluation points represented by red dots is 295.}
  \label{fig:simmap}
\end{figure}

\begin{figure}[tb]
  \centering
  \includegraphics[width=1.0\linewidth]{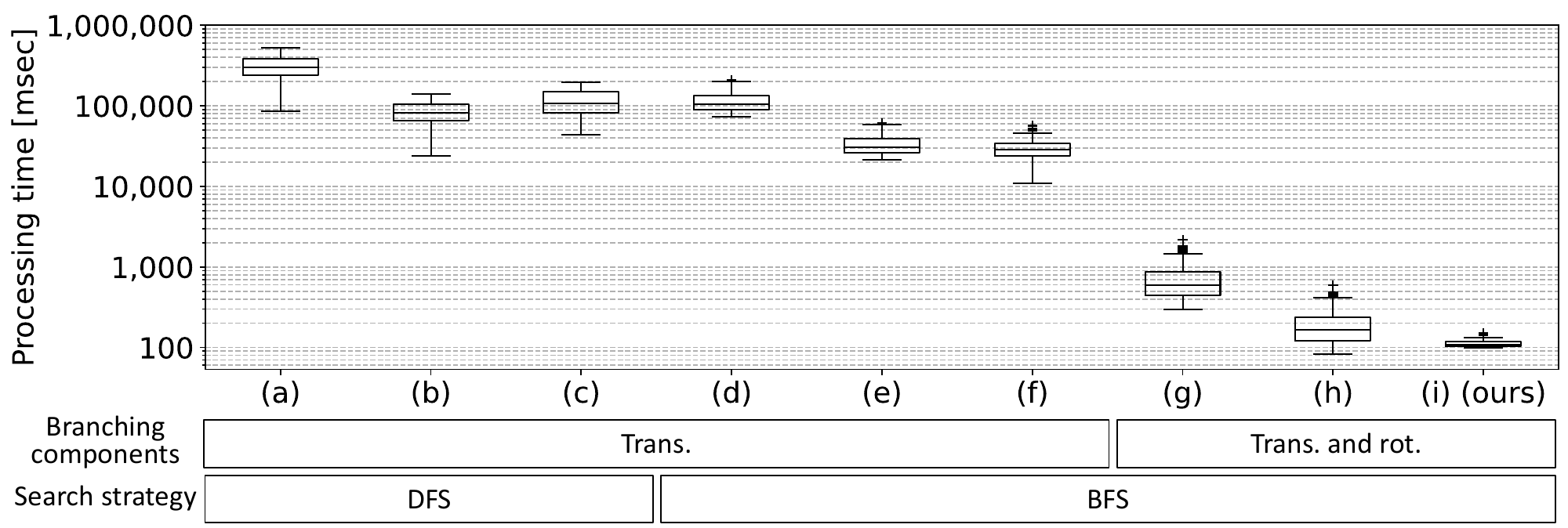}
  \caption{Processing time of the configurations (a) - (i) in the simulated environment.}
  \label{fig:sim-time}
\end{figure}

{\bf Experimental data:} To demonstrate the performance of the proposed algorithm, we created LiDAR data and ground truth sensor poses in a simulated environment, as shown in Fig. \ref{fig:awsim}. The 3D point cloud map data was captured in an urban environment containing many buildings, as shown in Fig. \ref{fig:simmap}. The map size was 211.0 $\times$ 729.7 $\times$ 68.5 $\rm{m^{3}}$. The total number of evaluation points was 295. We generated LiDAR scan point clouds with sensor parameters of the Velodyne VLP32C. We added random noise in the range of $[-0.01, 0.01]$ rad to the roll and pitch components of the LiDAR scan point clouds to imitate the gravity direction estimation errors. 

{\bf Results:} Configurations (a) - (i) successfully estimated the sensor poses with translation and rotation errors smaller than the criterion threshold values for all evaluation points. Along with the accuracy evaluation, we evaluated how often the proposed rotational branching showed incorrect upper bound estimates and how much the incorrect estimates exceeded the true upper bound. We confirmed that the rotational branching showed incorrect upper bound estimates for only 0.001\% of the cases, and the average error with respect to the true upper bound was about 1.5\%. These results show that rotational branching introduces only a small approximation.

Fig. \ref{fig:sim-time} shows the processing times of all configurations. Configurations (d) - (f) with BFS were slightly faster than configurations (a) - (c) with DFS. Because BFS first branches the node with the highest provisional score, it can find the global solution more quickly when the scan points have a large overlap with the map (i.e., the score of the solution is close to the maximum score). However, it was necessary to calculate a large number of nodes across all levels. 
In configurations (g) - (i), the combination of the rotational branch and BFS contributed to a significant reduction in processing time. Configuration~(i) with batch processing reduced the processing time by more than 50 \% from the configuration (g) with CPU single thread. Configurations (g) and (h) with CPU processing also demonstrated fast processing within less than 1 sec because a high best score in the simulated environment allows for faster pruning.
\subsection{Real Environment}
\begin{figure}[tb]
  \centering
  \includegraphics[width=0.88\linewidth]{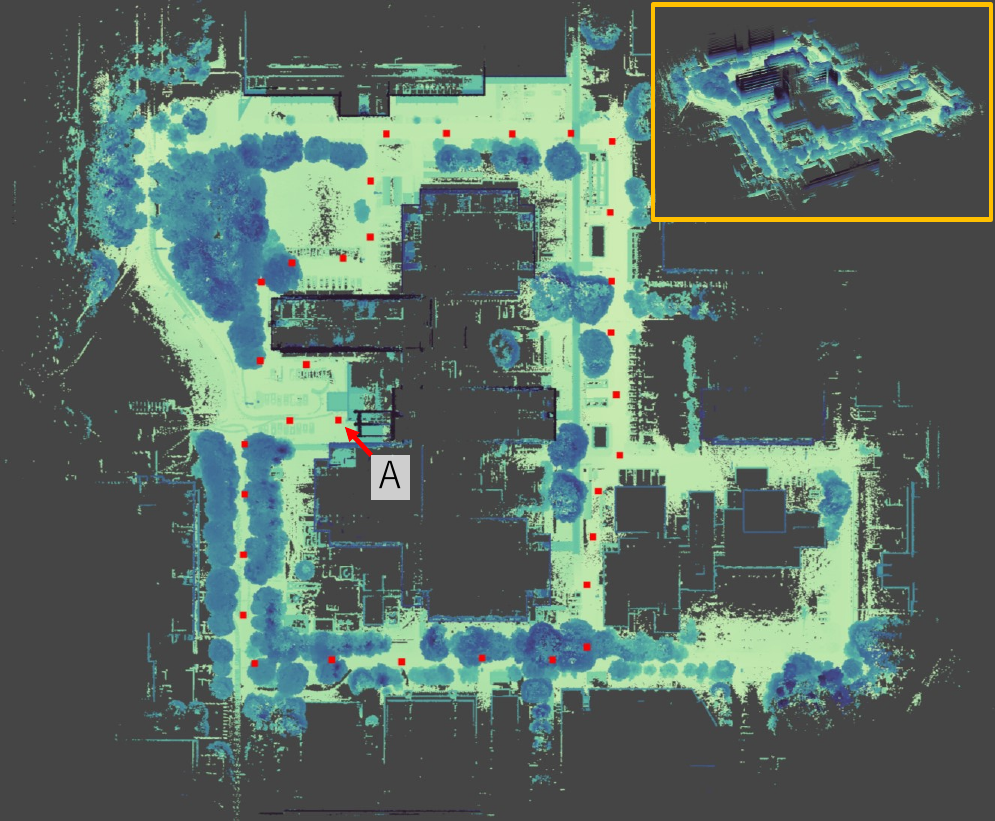}
  \caption{3D point cloud map in the experiment with real environment. The map size is 429.6 $\times$ 404.3 $\times$ 69.6 $\rm{m^{3}}$. The total number of evaluation points represented by red dots is 32.}
  \label{fig:aistmap}
\end{figure}

\begin{table}[tb]
  \centering
  \caption{Success rate and processing time}
  \label{tab:conventional}
  \begin{tabular}{g|ggg}
    \toprule
    \rowcolor{white}  Method & Trans. & Trans. and rot. & Processing time [msec] \\
	\midrule
    \rowcolor{white} TEASER++ \cite{teaser} & 16/32 & 11/32 & 45,573$\pm$14,462 \\
    Quatro \cite{quatro} & 18/32 & 18/32 & 42,047$\pm$14,038 \\
    \rowcolor{white} FGR \cite{fgr} & 17/32 & 14/32 & 43,677$\pm$15,283 \\
    Conf. (i) (ours) & \bf{32/32} & \bf{32/32} & \bf{878}$\pm$\bf{1,128} \\ 
    \bottomrule
  \end{tabular}
\end{table}

\begin{figure}[t]
  \centering
  \includegraphics[width=1.0\linewidth]{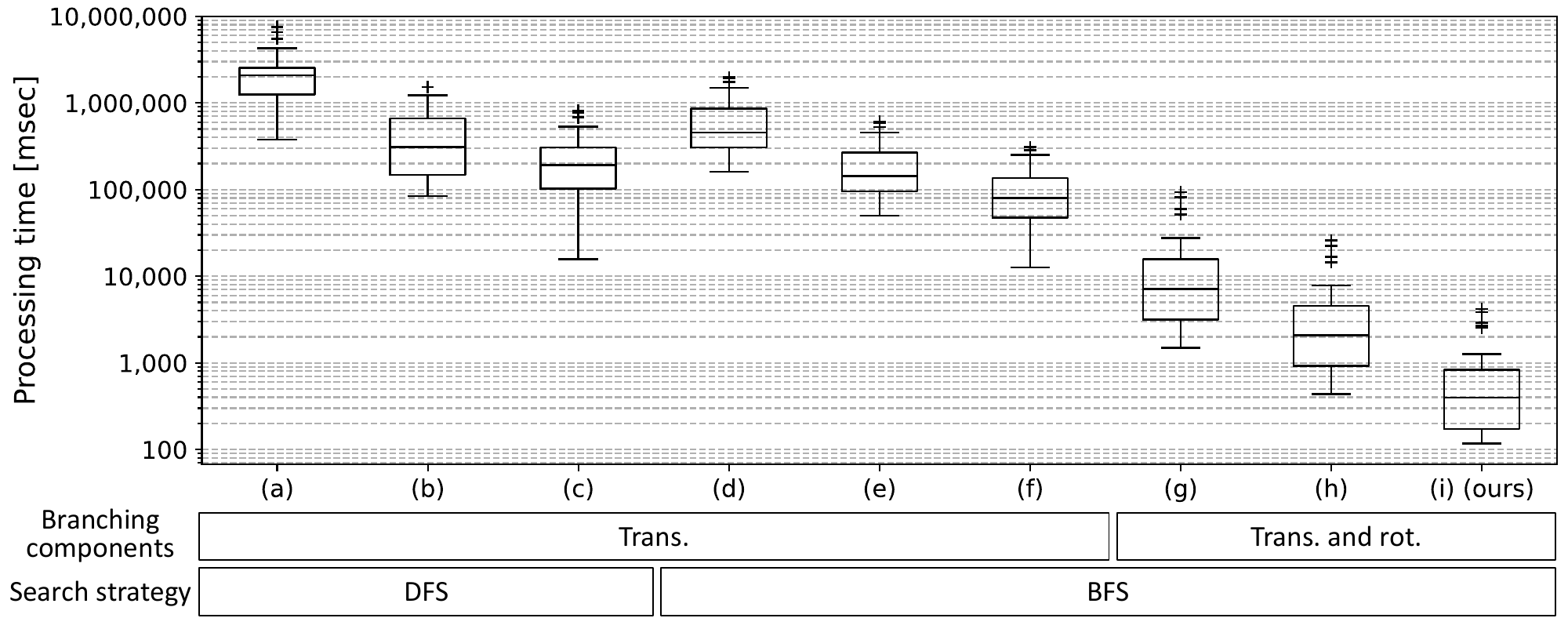}
  \caption{Processing time of the configurations (a) - (i) in the real environment.}
  \label{fig:aisttime}
\end{figure}

\begin{table}[t]
\centering
\caption{Processing breakdown [msec]}
\label{tab:breakdown}
\begin{tabular}{cl|r}
\toprule
\rowcolor{white}  Process & Procedure & \multicolumn{1}{c}{Average} \\
\midrule
\rowcolor{white}  \multirow{3}{*}{Preprocessing} & Create voxel maps & 9,272.2 $\pm$ 49.8 \\
& Set source point cloud & 30.5 $\pm$ 0.7 \\
\cmidrule(lr){2-3}
&Total & 9,302.7 $\pm$ 49.7 \\
\midrule
& Initial nodes calculation & 2.7 $\pm$  0.1 \\
& Find best score & 119.8 $\pm$ 1.4 \\
\rowcolor{white}  & Pop remaining queue & 50.9 $\pm$ 2.3 \\
\cmidrule(lr){2-3}
\rowcolor{white} \rowcolor{white}  \multirow{-4}{*}{Localization}  & Total & 173.4 $\pm$ 3.1 \\
\bottomrule
\end{tabular}
\end{table}

{\bf Experimental data:} 
We then compared the proposed method with several other state-of-the-art methods in a real environment. Fig. \ref{fig:aistmap} shows the outdoor point cloud map used in this experiment, containing buildings, parked vehicles, and street trees. The map size was 429.6 $\times$ 404.3 $\times$ 69.6 $\rm{m^{3}}$. The total number of evaluation points was 32. We used the Livox-Mid360 with a built-in IMU. We measured sensor data on a different day from when the map data was acquired, and thus the scan data had differences in parked vehicle positions and vegetation from the map data. The LiDAR scan was aligned in the direction of the average acceleration direction measured using the IMU. To obtain the ground truth sensor poses, we manually aligned each LiDAR scan with the map point cloud using local registration \cite{gicp}.

{\bf Estimation accuracy:} Table \ref{tab:conventional} summarizes the success counts and processing times of the evaluated global localization methods. The processing time was calculated only for successful results. Global registration methods \cite{teaser}, \cite{quatro},~\cite{fgr} succeeded at most 18 locations out of 32. Because there were many similar buildings in the map point cloud, these methods struggled with similar point features, which resulted in too many incorrect correspondences, and showed incorrect estimation results far from the ground truth locations several times.
In contrast, the proposed 3D-BBS successfully estimated the correct sensor locations for all evaluation data even under environmental changes, e.g., vegetation and dynamic obstacles.
Moreover, configurations (a) - (h) were also successful at all evaluation points in the real environment, similar to the simulation environment.

{\bf Processing time:} Fig. \ref{fig:aisttime} shows the processing times of all configurations. We observe that the proposed configuration~(i) performed the fastest processing in the real outdoor environment. Moreover, there was a large difference between configurations (g) and (h) with the CPU and configuration (i) with the GPU. This was because as the number of nodes to be branched increased, the overhead of the GPU batch processing got canceled, resulting in a large performance gap between the CPU and GPU implementations. However, unlike the simulated environment, the interquartile range was large overall. The more outliers included in the LiDAR scan, the more it affects the processing time. To further improve 3D-BBS performance, it is necessary to consider the inconsistency between the map and LiDAR scans. 

Table \ref{tab:breakdown} shows the breakdown of the processing time of 3D-BBS with the proposed configuration (i) at evaluation point A, as shown in Fig. \ref{fig:aistmap}. The preprocessing, which needed to be performed for the map only once, was completed within 10 sec, including the memory copy of the multi-resolution voxel map from the CPU to the GPU. The set source point includes the construction of a graph structure for an efficient parallel score calculation. In the localization step, the number of initial nodes was significantly reduced thanks to the rotational branching, and the initial node calculation took only 2.7 msec. Moreover, the branching operation accelerated by the batch process allowed us to find the best score in 119.8 msec. The localization process was completed by popping the remaining queue in 50.9 msec.
\section{Conclusion}
This paper proposed a 3D global localization approach based on BnB algorithm. We used a sparse hash table to overcome the memory increase in hierarchical 3D voxel maps. Moreover, we proposed roto-translational branching and batched processing to reduce the processing time. The experimental results showed that 3D-BBS accurately estimated the global pose and completed the localization process in a second.
In future work, we plan to extend the proposed method to handle extreme cases (e.g., degenerated areas or LiDAR scan occlusions).
\balance

\bibliographystyle{IEEEtran}
\bibliography{icra2024}

\end{document}